\documentclass[nonacm,sigconf]{acmart}
\setcopyright{none}
\renewcommand\footnotetextcopyrightpermission[1]{}
\begin{document}
	
\title{\texttt{EASTER}: Efficient and Scalable Text Recognizer}

\author{Kartik Chaudhary}
\affiliation{%
	\institution{Optum, UnitedHealth Group}
	\city{Bengaluru}
	\country{India}}
\email{kartik@optum.com}

\author{Raghav Bali}
\affiliation{%
	\institution{Optum, UnitedHealth Group}
	\city{Bengaluru}
	\country{India}}
\email{raghavbali@optum.com}

\renewcommand{\shortauthors}{Chaudhary and Bali, et al.}

\begin{abstract}
Recent progress in deep learning has led to the development of Optical Character Recognition (OCR) systems which perform remarkably well. Most research has been around recurrent networks \cite{8563238,doetsch,graves2009offline,graves2008novel,graves2006connectionist,pham2014dropout,poznanski,voigtlaender2016handwriting} as well as complex gated layers \cite{bluche2017gated,ingle2019scalable} and \cite{jianfeng2017deep} which make the overall solution complex and difficult to scale. In this paper, we present an \textbf{E}fficient \textbf{A}nd \textbf{S}calable \textbf{TE}xt \textbf{R}ecognizer (\texttt{EASTER}) to perform optical character recognition on both machine printed and handwritten text. Our model utilises 1-D convolutional layers without any recurrence which enables parallel training with considerably less volume of data. We experimented with multiple variations of our architecture and one of the smallest variant (depth and number of parameter wise) performs comparably to RNN based complex choices. Our 20-layered deepest variant outperforms RNN architectures with a good margin on benchmarking datasets like IIIT-5k and SVT. We also showcase improvements over the current best results on offline handwritten text recognition task. We also present data generation pipelines with augmentation setup to generate synthetic datasets for both handwritten and machine printed text.
\end{abstract}

\maketitle

\section{Introduction}
Text is a ubiquitous entity in natural images and most real world datasets like scanned documents, restaurant menu cards, receipts, tax forms, license plates, etc. These datasets may contain text in both, printed as well as handwritten formats. Extracting text information from such datasets is a complex task due to variety of writing styles and more so due to limitation of ground truth.
Optical Character Recognition (OCR) systems have been in existence for quite sometime now \cite{espana2010improving} and \cite{mori1999optical}. The improvements and research in Deep Learning, CNN and LSTM based OCR solutions \cite{8563238, doetsch, graves2009offline, graves2008novel, graves2006connectionist, pham2014dropout, poznanski, Borisyuk_2018} and \cite{voigtlaender2016handwriting} have taken the field by a storm. The results from these solutions are leaps and bounds ahead of traditional solutions like Tesseract\cite{smith2007overview}. The downside of these Deep Learning based solutions is their dependence on huge amounts of data and compute.

Handwritten Text Recognition or HTR is an even more involved process with countless variation of styles. While OCR for printed text has seen good improvements, HTR still remains a challenge. Lack of training data adds to the list of issues. Moreover, the models trained for printed text do not generalise well (even with transfer learning) onto HTR tasks.
 
\citeauthor{ingle2019scalable} in their work ``A scalable handwritten text recognition system''~ \cite{ingle2019scalable}, showcase improvements on limited datasets with architectures that have recurrent connections. They present Gated Recurrent Convolutional Layers (GRCLs) as specialised convolutional layers to perform recurrence along depth as compared to LSTMs which do the same along time dimension. In this paper we address the problem of training data volume and compute requirements together by presenting three variants of our fully convolution architecture devoid of recurrent connections. These models handle both, handwritten and machine printed texts. Being fully convolutional (using only 1-dimensional convolutions) enables development of smaller, faster and parallel trainable models. This further reduces the barrier for deployment and scalability. 

\citeauthor{coquenet2019} \cite{coquenet2019} challenge the notion of recurrent networks by presenting gated convolutional networks. They present results which are an improvement or are on par in comparison to CNN-BiLSTM based networks. Works by \citeauthor{DBLP:journals/corr/ShiBY15}\cite{DBLP:journals/corr/ShiBY15}, \citeauthor{jaderberg2014deep}\cite{jaderberg2014deep} and \citeauthor{DBLP:journals/corr/LeeO16}\cite{DBLP:journals/corr/LeeO16} improve further by utilising specialised attention mechanisms, complex recurrent convolutions and other enhancement techniques to solve the tasks of OCR and HTR.

Another major difference between \texttt{EASTER} and models described in \cite{ingle2019scalable, coquenet2019, bluche2017gated, Borisyuk_2018}, etc. (apart from architectural differences) is the training data. We focussed our model to perform OCR on word level inputs, i.e. each input is a single word. This restriction was based upon practical and deployment considerations of our application. This simplification also eases our dataset preparation and training process but does not limit the model's capability to handle line level inputs. We confirm our claims by showcasing performance on line level inputs as well. This shows the robustness of our model and its generalisation capabilities, all without using RNNs.

Our work is inspired by research in the field of Automatic Speech Recognition (ASR). Similar to text recognition, the task of speech recognition works upon a sequential input where label alignment is not a trivial task. ASR related work by \citeauthor{li2019jasper} and \citeauthor{DBLP:journals/corr/CollobertPS16} rely on using non-recurrent architectures. These works rely on multiple repeating block structures composed of different subcomponents. They also utilise residual connections and experiment with different depths to achieve state of the art performance. 

\begin{figure*}
	\begin{center}
	\includegraphics[width=0.9\linewidth]{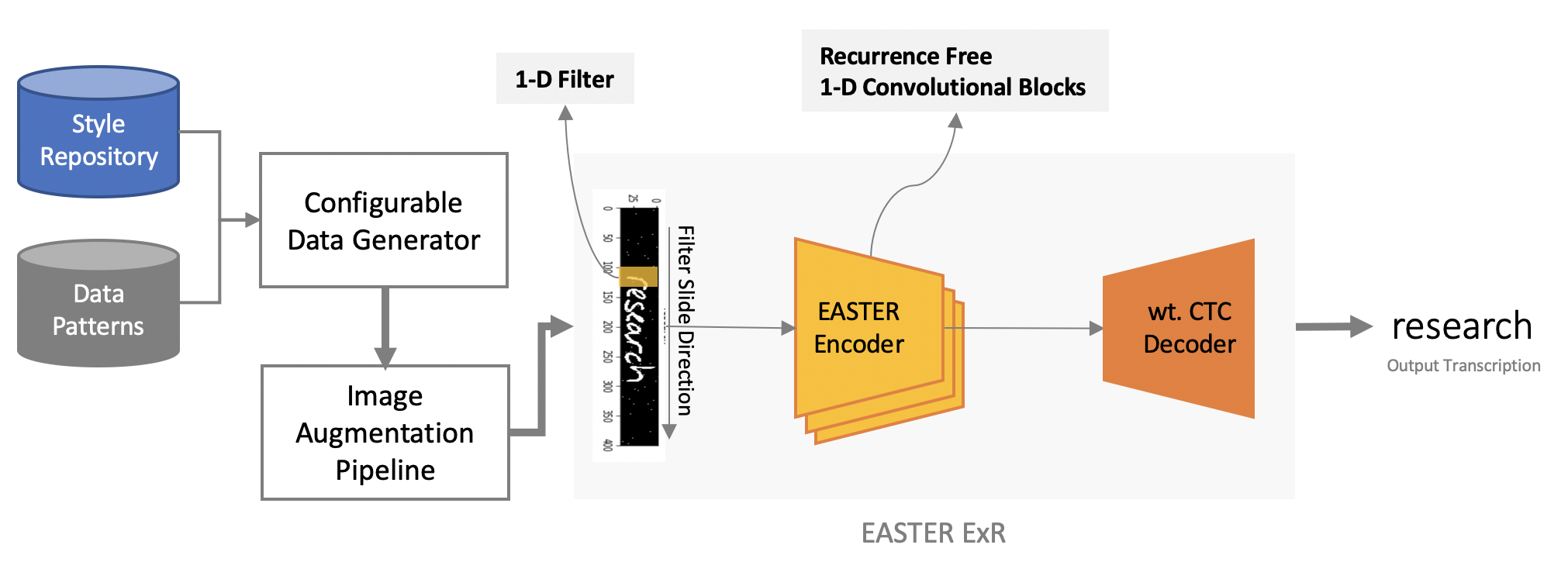}
	\caption{EASTER Pipeline consists of a configurable data generator, an image augmentation pipeline and recurrence free EASTER architecture. Given a list of data patterns like names, addresses, phone numbers, etc. along with a list of fonts, styles, etc. the configurable data generator can generate training datasets of required size. Augmentation pipeline helps add perturbations to generate realistic samples}
	\label{fig: overall}
	\end{center}
\end{figure*}

To the best of our knowledge, this is the first work that leverages only 1-D convolutions for the task of offline handwriting recognition without any recurrence. \texttt{EASTER} being a simple architecture, outperforms more complex models in terms of training time, volume of training data and performance (word and character error rates). Figure \ref{fig: overall} presents the overall setup.

The remainder of this paper is organised as follows:
Section 2 describes how we prepare different datasets to train \texttt{EASTER} for the tasks of OCR on machine printed and HTR on handwritten texts. We also outline the image augmentation pipeline to infuse variance in the training dataset. Section 3 discusses \texttt{EASTER} architecture in detail. In section 4 and 5 we discuss different experiments with variations in the \texttt{EASTER} architecture and present results across different datasets respectively. Section 6 concludes the paper.
\section{Data Preparation}
To prepare training data for both handwritten and machine printed text, we applied different methodologies. The task of preparing training data for handwritten text is a bit more difficult as compared to machine printed. First difficulty is the availability of annotated handwritten text followed by the large variations in handwriting styles. Manually preparing such datasets is time consuming and cost ineffective.

In the following subsections we will cover data preparation methodologies for both tasks in detail. This will be followed by details on augmentation of these datasets.

\subsection{Handwritten Text}
We utilised three different approaches to prepare the training dataset for the task of handwritten OCR. The IAM handwriting dataset \cite{Marti2002TheIA} contains stroke information for handwritten text collected from various contributors in an un-constrained settings. We leverage the images from the offline subset as input samples which have corresponding transcriptions available. This dataset has limited samples yet provides line level data points with different handwriting styles and text patterns.

\begin{figure*}
	\centering
	\includegraphics[width=0.9\linewidth]{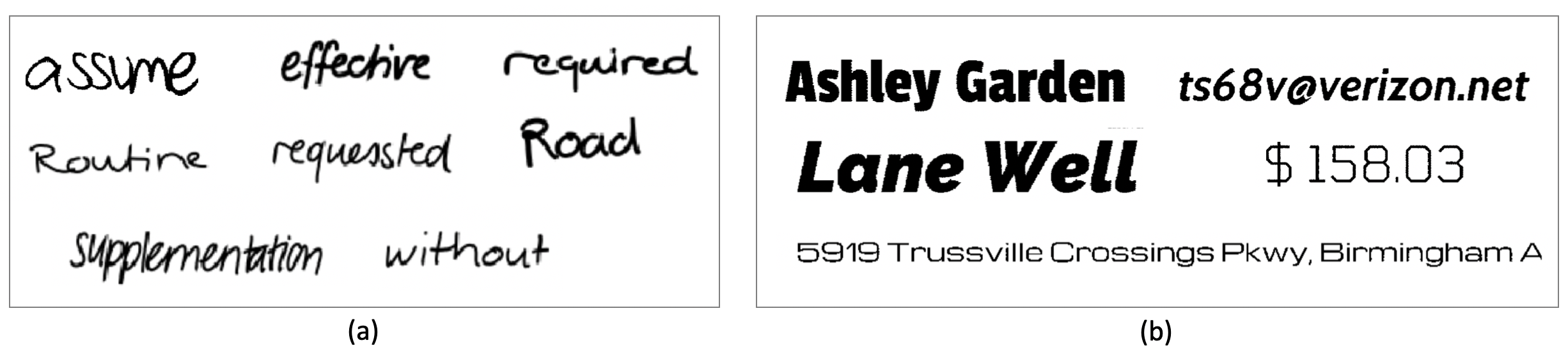}
	\caption{Synthetically Generated Random Samples; (a)Handwritten Text samples, (b) Machine Printed Text Samples for random name, email address, dollar value and street address  }
	\label{fig: machine_hand}
\end{figure*}

The next approach was to synthetically generate handwritten text data, similar to how we prepared the machine printed dataset. For this we leveraged the method presented by Alex Graves \cite{graves2013generating}  to generate handwritten text. Similar to GRCL \cite{ingle2019scalable}, we enabled the model to generate different styles and variations as well. We used specific hyper-parameters to control the mixture density layer and sampling temperature to generate usable samples.
The final approach was to manually write as well as label the samples to make sure we cover a wide variety of text patterns and style. See figure \ref{fig: machine_hand} for samples from our handwritten text dataset.

Together these three approaches led to the training dataset we utilised for training our HTR model. Note that for some experiments we utilised only specific subsets of this dataset, more details in the experiments section. 

\subsection{Machine Printed Text}
Collecting machine printed dataset from with-in the organisation and from different public sources has certain restrictions and issues. These factors range from datasets being very clean, limited fonts or having only specific type of patterns. Instead of collecting and curating data from such sources, we devised an ingenious method for synthetically generating a dataset for machine printed text. 


The first step was to assemble a list of common patterns of text in the real world. These include text for street addresses, names, dollar amounts, etc.. The second step was to prepare a repository of different font styles, strokes, formatting (underline, italics, etc.). The final step was to use the patterns and styles as inputs to a probabilistic synthetic text data generator. The generator was designed to have configurable settings to adjust the variation in styles, patterns, length of text samples and overall dataset size. See figure  \ref{fig: machine_hand} for samples from our machine printed text dataset. Using this method, we virtually have the ability to generate infinite such datasets for training and improving our models.

\subsection{Augmentation}
Text available in real world is rarely in noise free conditions. Issues such as bad scan quality, contrast issues, faded ink (especially in case of handwritten text), etc. complicate the tasks of OCR and HTR even more. To develop models which can handle such issues, we added certain augmentations to our datasets. The augmentation pipeline was designed to add perturbations to our datasets which would help mimic real world scenarios. Inspired by degradation pipeline in \cite{ingle2019scalable} and augmentation library imgaug \cite{imgaug} we added augmentations to both, machine printed and handwritten training datasets. Noisy backgrounds are a common feature, hence gaussian noise, salt and pepper, fog, speckle, random lines/strokes, etc. we used. To handle different sizes, we made use of padding across the four edges or a combination of two to three edges to generate different sizes. Perspective related augmentations involved rotation, warping, dilation, shear, etc.  

\begin{figure}
	\centering
	\includegraphics[width=0.9\linewidth]{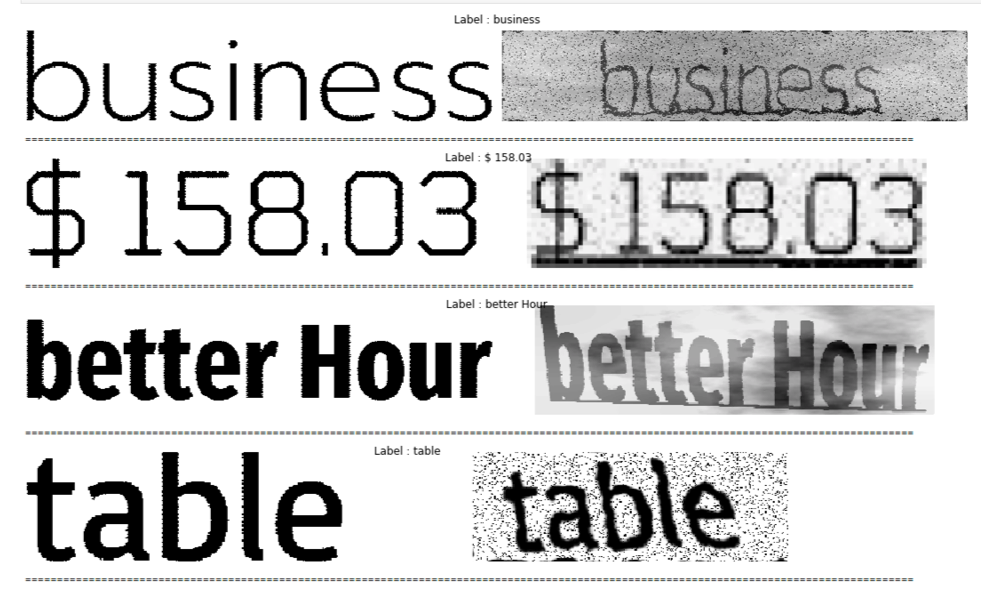}
	\caption{Augmented Samples. Each row represents a random sample from the synthetic dataset where left image is the initial output and the right image represents the output after application of one or several augmentations.}
	\label{fig: augmentations}
\end{figure}

These augmentation techniques were applied to our datasets using a probabilistic pipeline to ensure enough variability and representation. See figure  \ref{fig: augmentations} for sample augmentations.

\section{Architecture}
OCR and HTR both involve taking images with text as input and generating corresponding text as output. Recurrent architectures make use of LSTMs to capture sequential dependency followed by Connectionist Temporal Loss (CTC)\cite{graves2006connectionist} to help train models without specifically aligning inputs and their labels. 

The basic architecture we present here is a 1-D Convolutional network inspired by the research in the field of Automatic Speech Recognition (ASR) tasks. Works of \citeauthor{li2019jasper} \cite{li2019jasper} \citeauthor{DBLP:journals/corr/CollobertPS16} \cite{DBLP:journals/corr/CollobertPS16} and \citeauthor{DBLP:journals/corr/abs-1812-07625} \cite{DBLP:journals/corr/abs-1812-07625} highlight the effectiveness of convolutional networks in handling sequence to sequence tasks (ASR) without recurrent connections. We extend the similar thought process for the field of OCR and HTR using \texttt{EASTER} (see figure \ref{fig: EASTER_ARCH}).



\subsection{\texttt{EASTER} Encoder}
\begin{figure}
	\centering
	\includegraphics[width=0.9\linewidth]{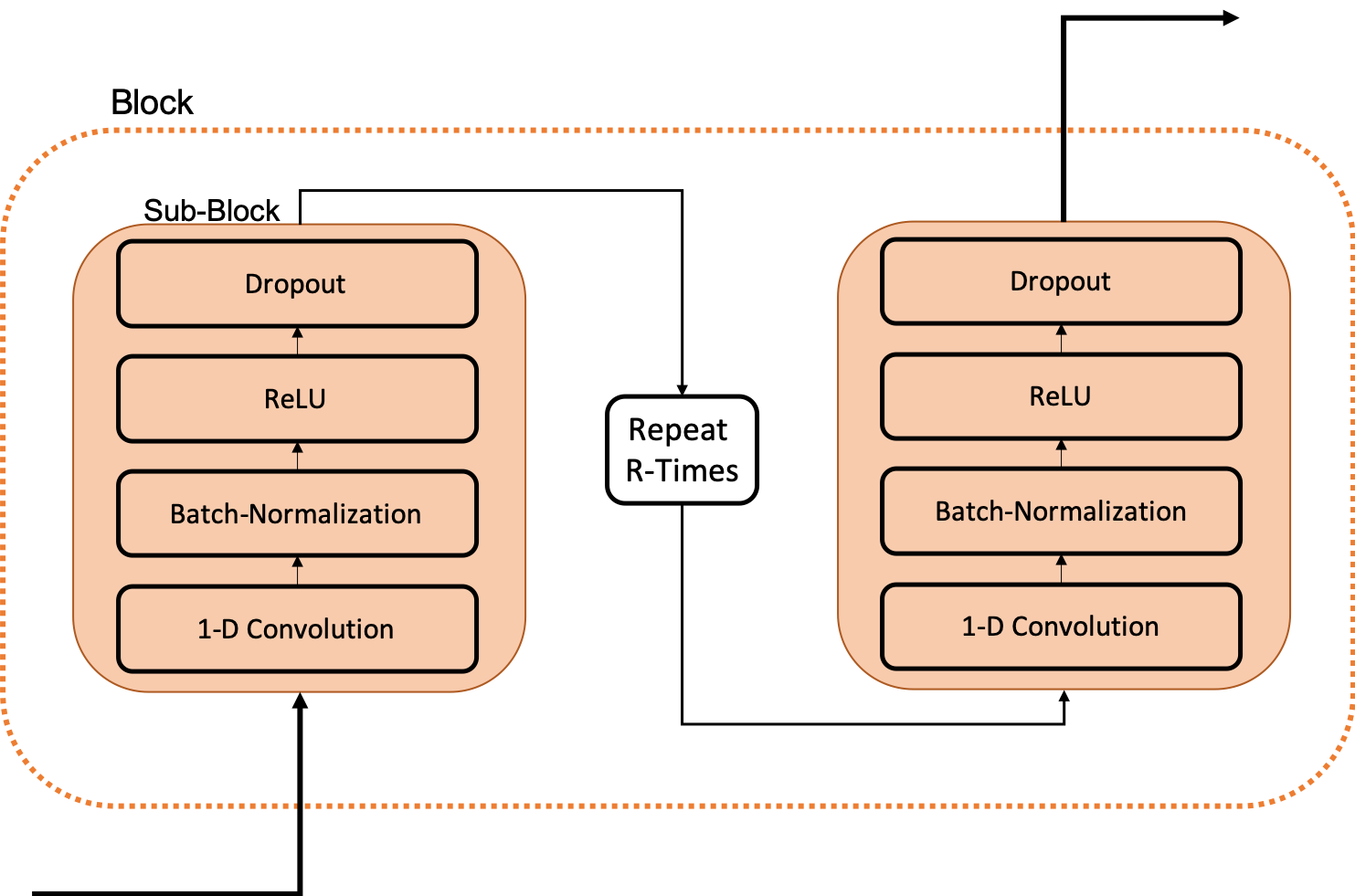}
	\caption{Components of \texttt{EASTER} block. Each block contains multiple repeating sub-blocks consisting of layers for 1-D Conv, Batch Normalisation, ReLU and Dropout. Different blocks utilize different number of convolutional filters and other hyperparameters.}
	\label{fig: easter_block}
\end{figure}

\texttt{EASTER} follows a block approach where-in each block consists of multiple repeating sub-blocks. Each sub-block comprises of a 1-D Convolutional layer with multiple filters followed by layers for normalisation, ReLU and dropout. We utilise padding to maintain the dimensions of the input slice. Each \texttt{EASTER} architecture has 1 preprocessing block and 3 post-processing blocks. The pre and post processing blocks also follow similar block structure. In our experiments, we found Batch-Normalisation to outperform other normalisation techniques, this is similar to the findings mentioned in \cite{li2019jasper}. Figure \ref{fig: easter_block} shows a sample \texttt{EASTER} block. Table \ref{tab:eeaster33} shows the structure of a 3x3 EASTER model. The table outlines the number of blocks, sub-blocks, number of filters and other hyper-parameters.

\begin{center}
\begin{table*}[h]
\begin{tabular}{@{}lllllll@{}}
\hline
\multicolumn{1}{c}{Block \#} & \# of Sub-Blocks & Kernel & \# of Filters & Dropout & Dilation & Stride\\ 
\hline\hline
Preprocess-I             & 2 & 3 & 64      & 0.2 & 1 & 2 \\
B1             & 3 & 3 & 128     & 0.2 & 1 & 1 \\
B2             & 3 & 4 & 128     & 0.3 & 1 & 1 \\
B3             & 3 & 6 & 128     & 0.3 & 1 & 1 \\
Postprocess-I   & 1 & 7 & 256     & 0.4 & 2 & 1 \\
Postprocess-II  & 1 & 1 & 512     & 0.4 & 1 & 1 \\
Postprocess-III & 1 & 1 & |Vocab| & 0   & 1 & 1 \\ 
\hline
\end{tabular}
\caption{EASTER 3x3: 9 blocks each consisting of 3 1-D Convolutional sub-blocks, 1 preprocessing block and 3 post processing blocks. Overall the model contains 14 layers with 1Million trainable parameters.}
\label{tab:eeaster33}
\end{table*}
\end{center}

\subsection{CTC and Weighed CTC Decoder}
The Connectionist Temporal Classification (CTC) method \cite{graves2006connectionist} is used to train  as well as infer results from our models. The characters (or vocabulary for our task of OCR/HTR) in the input image vary in width and the spacing. CTC enables us to handle such a task without the need to align input images and ground truth. 

We denote the training dataset as $D = \{X,Y\}$, where $X$ is the input image for transcription and Y is the label or ground truth. Assuming we have a vocabulary set $L$, then $Y=L^s$, where $s$ represents label length. CTC generates outputs at every time step $t$. The final output contains a sequence of repeating consecutive characters with $\epsilon$  (denoting blank space) in between. Thus, we add another symbol $\epsilon$ representing a blank to the vocabulary set $L$.
 \begin{equation}
 L^+ = \{L\cup\epsilon\}
\end{equation}
The objective is to minimise the negative log probability of obtaining $Y$ given an input $X$, i.e.
\begin{equation}
Objective = -\Sigma_{(X,Y)\epsilon D} logp(Y|X)
\end{equation}
The final output is obtained by merging consecutive repeating characters delimited by $\epsilon$. For instance, an output sequence like $1{\epsilon}bb{\epsilon}{\epsilon}a$ maps to $1ba$. To obtain such an output, we define a function $\gamma$ which squeezes repeating characters to single occurrence and removes blanks ($\epsilon$). Thus, $\gamma$ is a function which maps the intermediate repeating sequence output (denoted as $\pi$) to the final output $y$.
\begin{equation}
\label{eq:3}
p(y|X) = \Sigma_{\gamma(\pi)=y}logp(\pi|X)
\end{equation}
\begin{equation}
p(\pi|X) = \Pi_{t=1}^Ty_{\pi_t}^t
\end{equation}
, where $y_{\pi_t}^t$ is the probability of generating label $\pi_t$ at time $t$. 
Thus, the predicted label $y$ for input $X$ is given as:
\begin{equation}
y = \gamma(argmax_{\pi}p(\pi|X))
\end{equation}

\begin{figure*}
	\begin{center}
	\includegraphics[width=0.9\linewidth]{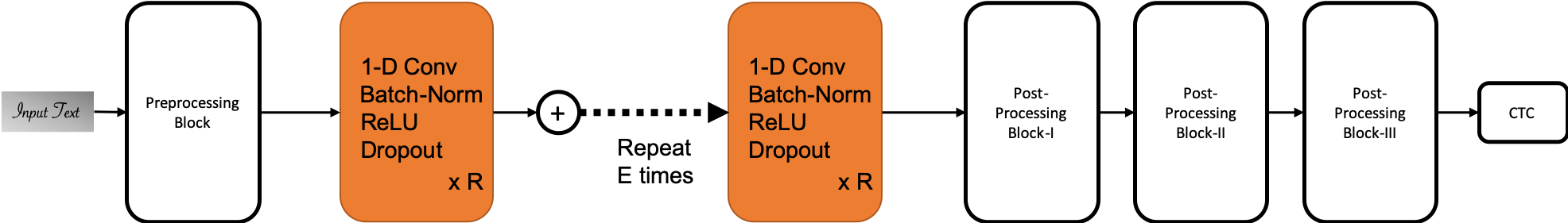}
	\caption{Basic ExR \texttt{EASTER} Architecture with 1 preprocessing block and 3 post-processing blocks}
	\label{fig: EASTER_ARCH}
\end{center}
\end{figure*}

Due to the way this function is designed, the model generates far too many $\epsilon$'s as compared to actual characters. This leads to the model being biased towards the blank class ($\epsilon$). To address this problem, \citeauthor{li2019novel} \cite{li2019novel} show multiple ways of adjusting the class weights for CTC. These weighting strategies address the problem of class imbalance and result in fast convergence. The class weighted CTC method is denoted as:
\begin{equation}
Class\_Weighted\_CTC(y|X) = -\Sigma_{t}\Sigma_{k}\alpha_{k}y_{k}^{t}logy_k^t
\end{equation}
, where $y_{k}^{t}$ is the generated output at time $t$ and,

\begin{equation}
\label{wams} \alpha_k =
\begin{cases}
  1-\alpha    & \text{if k}=\epsilon\\
  \alpha & \text{otherwise}
  \end{cases}
\end{equation}
In our experiments we saw significant improvement in performance with weighted CTC while training on small datasets. The most basic architecture for \texttt{EASTER} is shown in figure \ref{fig: EASTER_ARCH}. It is a 3x3 architecture with separate preprocessing and post-processing blocks. 

\subsection{Training}
For a typical training input, we first transform the input image into grayscale followed by scaling it down to a height of 40-pixels. The network is able to handle variable width inputs and requires no additional transformations. Individual characters have specific local structures and we utilise overlapping 1-D convolutions to exploit the same. 1-D Convolutions also capture short term sequential dependencies across sliding frames and further assist in capturing the underlying temporal aspects of the sequence. \texttt{EASTER} block architecture enables it to learn higher level features without the need of recurrence or any specialised gated mechanism. We trained our models using Keras \cite{chollet2015keras} with Tensorflow \cite{tensorflow2015-whitepaper} backend.

The smallest model with half million (0.5M) parameters achieves best results in under 1 hour of training.This enables faster turn-around time for retraining, experiments and ease of deployment. We experimented with two more variants , one with increased number of filters and the other with more depth. The deepest variant has 28M parameters and outperforms RNN based models with a good margin on our internal dataset.

\section{Experiments and Results}

\subsection{Handwriting Task}
We evaluate \texttt{EASTER}'s performance on IAM dataset for the task for HTR. To enable comparison with GRCL \cite{ingle2019scalable}, we follow the same process of training our models on different combinations of the IAM dataset. \citeauthor{ingle2019scalable} firstly train GRCL with IAM-offline dataset which contains only 6161 samples and report results on the IAM-offline test dataset. We repeat the same experiment with \texttt{EASTER}-5x3 (20 layer variant) and showcase improvements on both WER and CER by a good margin. We also went a step ahead and trained our model on only first 3000 samples out of 6161 samples ,i.e. only 50\% of the actual training data. \texttt{EASTER}-5x3's performance on 3k samples was observed to be better than GRCL with 6161 training samples.

\begin{table*}
\begin{center}
\begin{tabular}{@{}llllll@{}}
\hline
\multicolumn{1}{c}{Model} &
  \multicolumn{1}{c}{Training Dataset} &
  \multicolumn{1}{c}{\begin{tabular}[c]{@{}c@{}}\# of \\ Training Samples\end{tabular}} &
  \multicolumn{1}{c}{Augmentation} &
  \multicolumn{1}{c}{WER} &
  \multicolumn{1}{c}{CER} \\ 
 \hline\hline
\texttt{EASTER} & IAM-Off                        & \textbf{3,000}           & No  & \textbf{33.3} & \textbf{14.0} \\
GRCL   & IAM-Off                        & 6,161           & No  & 35.2          & 14.1         \\
\texttt{EASTER} & IAM-Off                        & 6,161           & No  & \textbf{25.4} & \textbf{9.8} \\ \hline
GRCL   & IAM-Off + IAM-On-Long & 511,524         & No  & 22.3          & 8.8          \\ 
\texttt{EASTER} & IAM-Off + IAM-On-Long & \textbf{24,481} & No  & \textbf{20.6} & \textbf{7.9} \\
GRCL   & IAM-Off + IAM-On-Long & 511,524         & Yes & 17.0          & 6.7          \\ 
\hline
\end{tabular}
\end{center}
\caption{Results on IAM Offline Dataset as measured using Word Error Rate (WER) and Character Error Rate (CER). GRCL refers to Gated Recurrent Convolutional Layers as presented by \citeauthor{ingle2019scalable} \cite{ingle2019scalable}}
\label{tab:htr_results}
\end{table*}

In the second experiment performed by \citeauthor{ingle2019scalable}, they concatenated smaller input samples in different combination to form a larger training dataset. The resulting training dataset has 511,524 training samples. This is a significantly larger training dataset as compared to the first experiment. We performed this experiment without the augmentation pipeline and observe a better performance from \texttt{EASTER}. Table \ref{tab:htr_results} refers to the two experiments. 

We observed similar performance boost on internal datasets. The improvements in WER and CER along with need for lesser training data and a lighter model (in terms of trainable parameters and memory footprint) helped us meet production requirements as well.

\subsection{Machine Printed Tasks}
We performed multiple separate experiments using \texttt{EASTER} for the task for OCR as well. The first experiment was based on the dataset prepared using our data generation pipeline. We utilised this dataset to benchmark our performance for internal datasets/use cases. 

The second set of experiments involved preparing our model for performance on some of the benchmarking datasets for text extraction from natural images. Popular benchmarking datasets we used for our experiments are as follows:
\begin{itemize}
  \item \textbf{IIIT-5k} \cite{MishraBMVC12}: This dataset consists of 2000 training images collected from the internet. There are about 3000 test samples. The dataset also consists of 50 and 100 word lexicon which we do not use during our experiments.
  \item \textbf{Google-SVT}  \cite{wang2011end}: is another interesting dataset consisting of only 257 training images and 647 images for test. Unlike \citeauthor{jianfeng2017deep}, we do not use the lexicon for our experiments.
\end{itemize}

The models with best results on the benchmark datasets utilise a range of techniques to train. Since the volume of training data in these datasets is not enough,  
\textbf{Synth-90k} \cite{jaderberg2014synthetic} dataset is used as a training set. As the name implies, this dataset generates realistic synthetic images. It consists of 900k test images and about 7million for training. Other techniques like preprocessing input crops to handle skew, rotation, contrast, super-resolution etc. are also applied by some of the works.
For our benchmarking experiments we only make use of \textbf{Synth-90k} dataset for training our models. We do not apply any preprocessing techniques other than resizing. Our experimental setup consists of a 20 layer deep variant, i.e. \texttt{EASTER}-5x3), model with residual connections. This model has 28million trainable parameters and a vocabulary of 62 characters from the English alphabet (26 lower-case + 26 Upper-case + 10 numerals). 

We did not apply any augmentations to our training dataset and performed greedy decoding. No  additional post processing steps like usage of language models etc were applied. There were two major reasons behind experimenting without a language model. Firstly, training language model requires additional time and data which is not always available. Secondly, and more importantly, the usage of language model slows down the inference pipeline in practice. Since our aim was to develop a deployable and usable OCR/HTR model, we were experimented without a language model for \texttt{EASTER}. 

Compared to \cite{DBLP:journals/corr/ShiBY15, jaderberg2014deep, DBLP:journals/corr/LeeO16},  \texttt{EASTER} comfortably improves upon the WER and CER performance against the \textbf{IIIT-5k} dataset. Our model is able to achieve 86.76\% word accuracy with 4.56\% character error rate. For case of \textbf{Google-SVT}, our model achieves a word accuracy of 78.51\% with a 9.7\% character error rate. We did not observe any improvements while using the training images from \textbf{Google-SVT} for fine-tuning. Our model improves upon the best in case of \textbf{IIIT-5k} while nearly achieves benchmark results on \textbf{Google-SVT}. It is important to note that in both cases, the inference was done with a greedy decoder without relying on the lexicon. The results and comparison are shown in detail in table  \ref{tab:ocr_results}  for reference.

\begin{table}
\begin{center}
\begin{tabular}{@{}lll@{}}
\hline
\multicolumn{1}{c}{Model}                                  & SVT    & IIIT-5k \\ 
\hline \hline
\citeauthor{jaderberg2014deep}            & 71.1\% & -       \\
\citeauthor{DBLP:journals/corr/LeeO16}    & 80.7\% & 78.4\%  \\
\citeauthor{DBLP:journals/corr/ShiBY15}   & 80.8\% & 78.2\%  \\
\citeauthor{jianfeng2017deep}             & \textbf{81.5\%} & 80.8\%  \\ \hline
\textbf{\texttt{EASTER}}-5x3 & 78.5\% & \textbf{86.76\%} \\ \hline
\end{tabular}
\end{center}
\caption{Machine printed natural text recognition word accuracies. All results are in unconstrained setting, i.e. models do not use lexicons during decoding. \texttt{EASTER} 5x3 in particular uses only greedy decoding in contrast to others in the table}
\label{tab:ocr_results}
\end{table}

\begin{figure*}
	\begin{center}
		\includegraphics[width=0.9\linewidth]{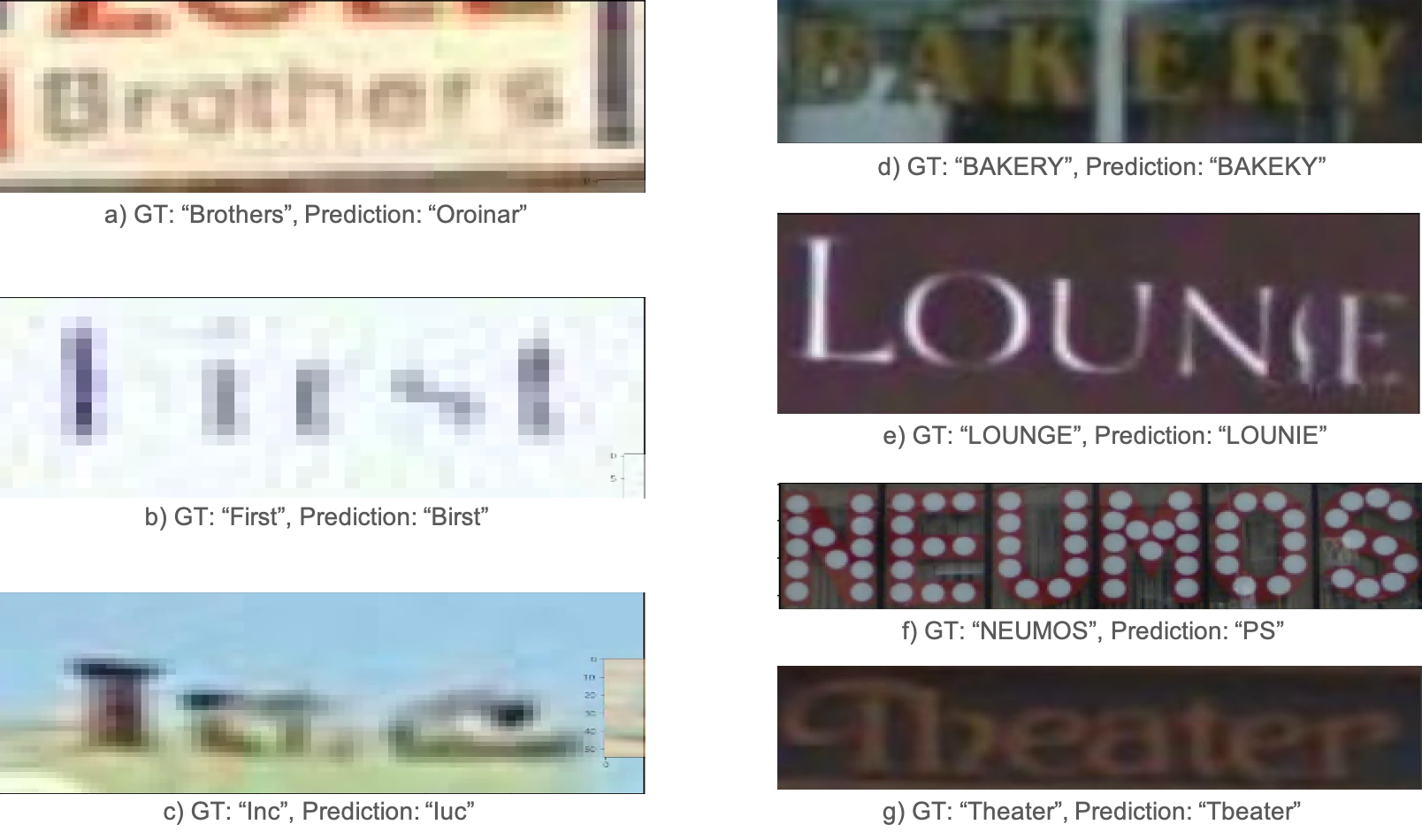}
		\caption{Samples where \texttt{EASTER} fails to transcribe correctly. Each example consists of ground truth followed by model output. Samples (b) and (e) can be attributed to distortions while samples (f) and (g) are associated with font related issues.)}
		\label{fig: failure}
	\end{center}
\end{figure*}

\begin{figure*}
	\begin{center}
		\includegraphics[width=0.9\linewidth]{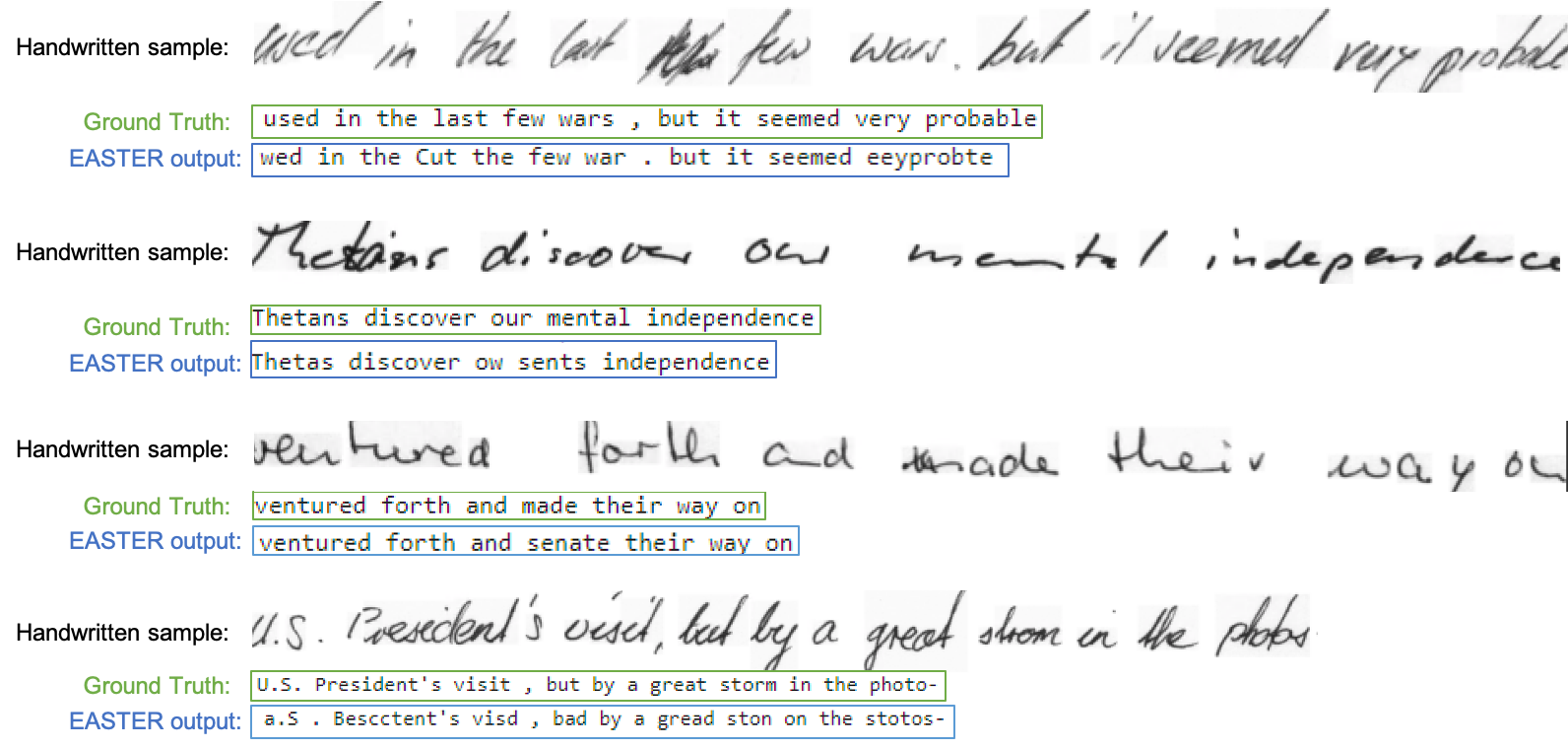}
		\caption{\texttt{EASTER} for handwritten text. Samples showcase scenarios where overlapping and highly cursive handwriting lead to incorrect transcriptions.}
		\label{fig: hw_failure}
	\end{center}
\end{figure*}

\begin{figure}
	\begin{center}
		\includegraphics[width=0.9\linewidth]{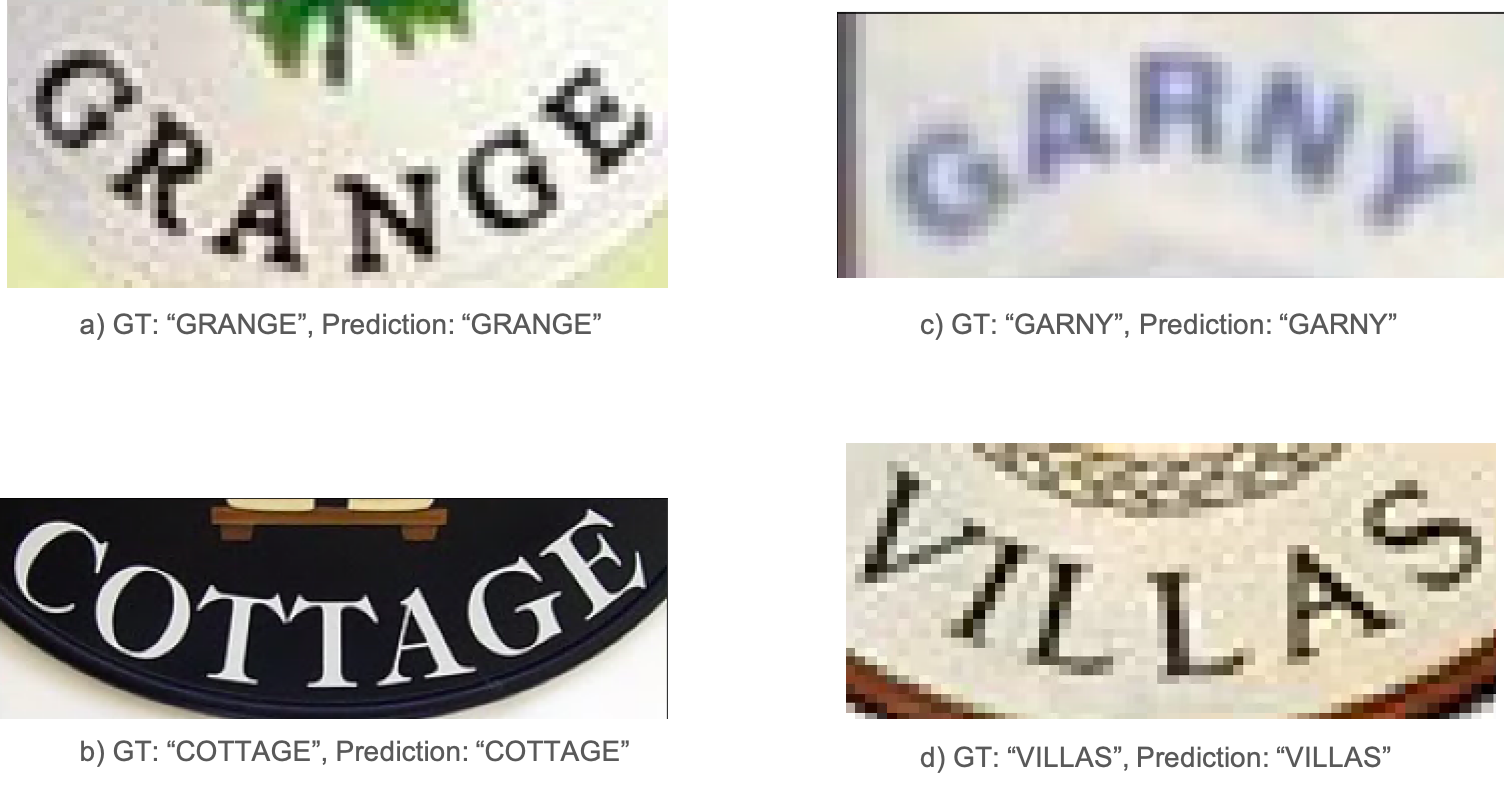}
		\caption{\texttt{EASTER} is a robust model which handles crops which contain curved text as well. Each example consists of ground truth followed by model output}
		\label{fig: curve}
	\end{center}
\end{figure}

\section{Limitations and Discussion}
Although our work can achieve compelling results in many cases, the results are not so favourable in some scenarios. Figure \ref{fig: failure} highlights some of the failure modes for IIIT-5k and Google-SVT datasets. We observed that even though the transcriptions do not match ground truth, the outputs seem to be honest mistakes. These mistakes can be largely attributed to issues like bad image quality along with distortions which are completely transforming certain alphabets. For instance, consider the distortions of alphabets "f" and "G" in figure \ref{fig: failure}(b) and  figure \ref{fig: failure}(e) respectively. The alphabets have been distorted to such an extent that it is nearly impossible to pick the correct alphabet without extensive post-processing. There are also cases where specific fonts seem to confuse the model (see figure \ref{fig: failure}(f) and  \ref{fig: failure}(g)). While distortions are tricky to handle, font related issues can be tackled using more training examples. 

The handwritten text recognition task is slightly more complex than the machine printed case. Though \texttt{EASTER} outperforms benchmarks both in terms of performance as well as training and compute requirements, it does face challenges when presented with overlapping and highly cursive handwritings. Figure \ref{fig: hw_failure} presents a few failure modes. Most issues in the highlighted examples can be attributed to unintelligible scribbles or at times hard to distinguish shapes. For instance, in the second example in figure \ref{fig: hw_failure}, the word "our" is misread as "ow" which is again a good enough guess given that the model is not making use of any language model or other post-processing steps.

It is also important that we present cases which highlight the robustness of our setup. The \texttt{EASTER} setup was largely trained on straight crops with a few augmentations catering to rotation and skew of overall text. Despite not being explicitly trained on curved crops, the model is able to handle such cases with ease. Figure \ref{fig: curve} presents a few such examples of successful transcription of curved texts.

Additional samples to visualize model performance on machine printed and handwritten text are available in the appendix section.

\section{Conclusion and Future Work}
We presented a fast, scalable and recurrence free architecture called \texttt{EASTER} for handwritten and machine printed text recognition tasks. Inspired from fully convolutional architectures for automatic speech recognition, we discussed the building blocks and training process for \texttt{EASTER}. We described the dataset preparation process for both handwritten and machine printed text along with a data augmentation pipeline. We also discussed about the impact of weighted CTC towards faster convergence. We finally presented results on benchmarking datasets for both HTR and OCR tasks. \texttt{EASTER} is able to achieve significant improvements in WER and CER. \texttt{EASTER}'s performance on internal datasets achieved near state of the art performance. We also showcased that the model performs equally well on line level data. Due to lesser number of trainable parameters and eventually smaller memory footprint, \texttt{EASTER} is easier to train and faster to use in production applications without much tooling. 

As part of future work we plan to utilise attention mechanisms to improve the decoding stage. We also plan to look into quantisation techniques to further reduce the memory footprint.

\section*{Acknowledgments}
We would like to thank Kishore V Ayyadevara and Yeshwanth Reddy for their work and contributions to the OCR project and making sure it gets widespread adoption, Vineet Shukla for helpful discussions and inputs to improve the solution, and the whole OCR team for their contributions.

\newpage
\bibliographystyle{ACM-Reference-Format}
\bibliography{references}

\begin{figure*}
	\begin{center}
		\includegraphics[width=0.9\linewidth]{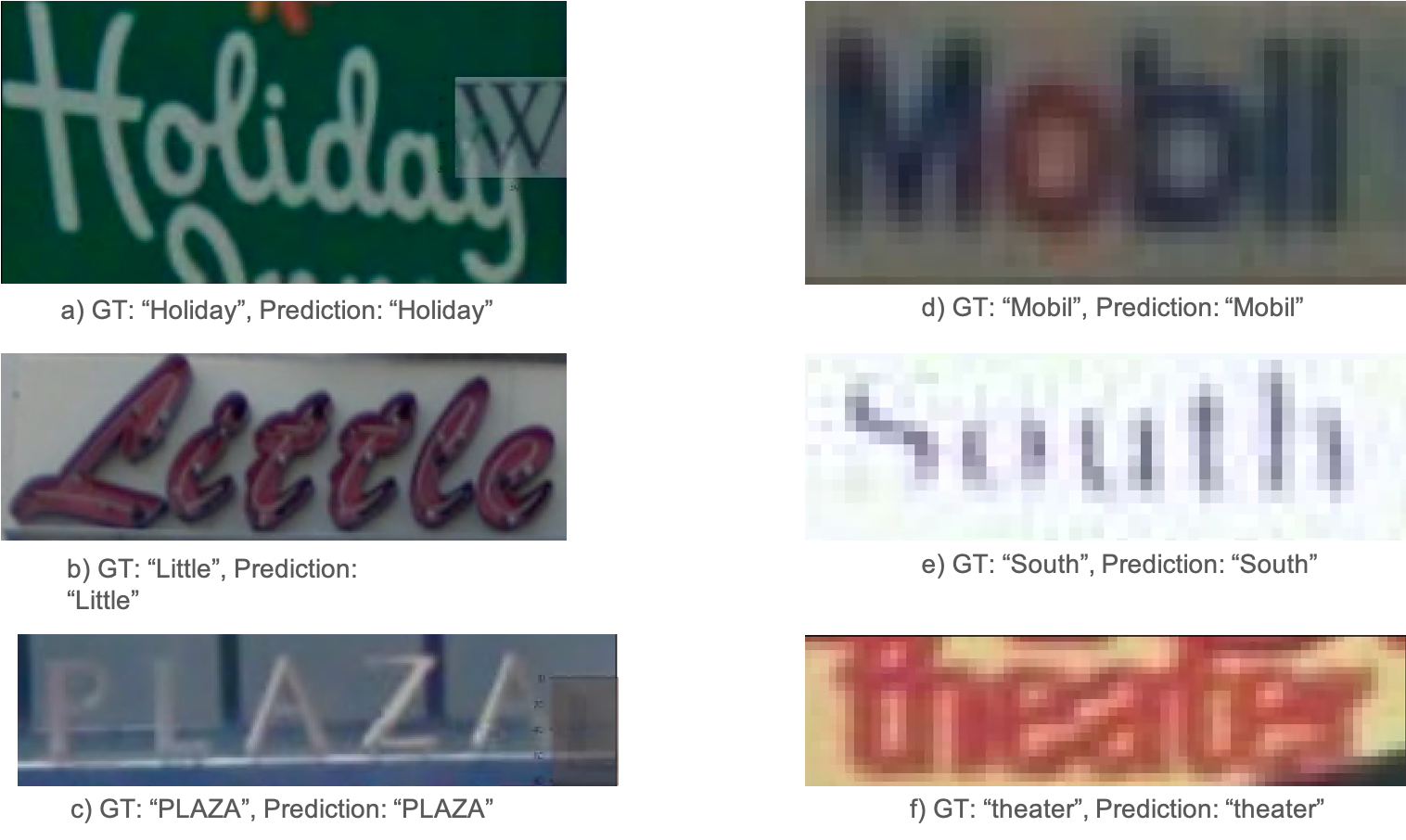}
		\caption{\texttt{EASTER} for Machine Printed Text. Despite unclear images and distortions, transcriptions are of high quality. Each example consists of ground truth followed by model output}
		\label{fig: machinie_success}
	\end{center}
\end{figure*}
\begin{figure*}
	\begin{center}
		\includegraphics[width=0.9\linewidth]{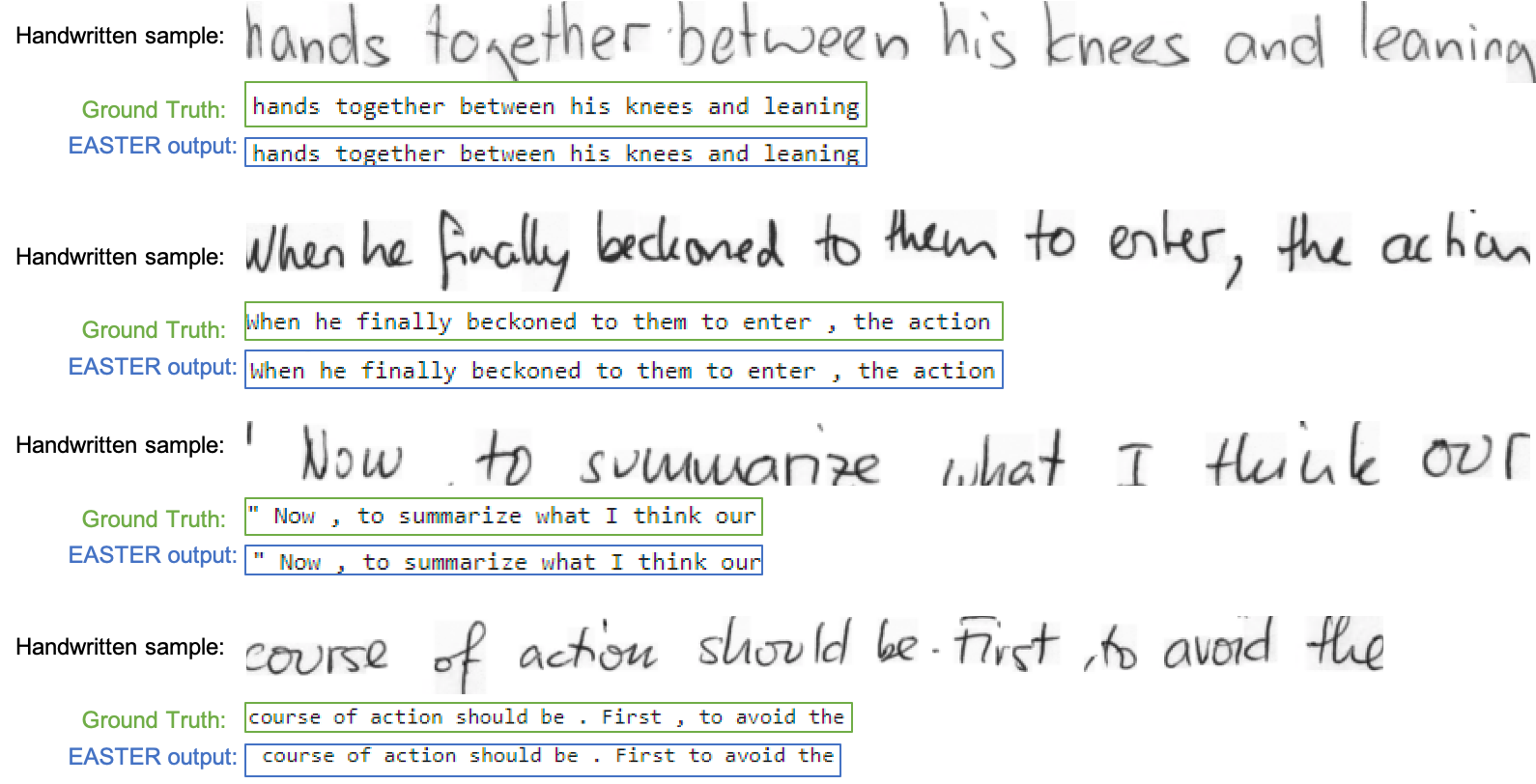}
		\caption{\texttt{EASTER} for Handwritten Text. Each example consists of ground truth followed by model output}
		\label{fig: hw_success}
	\end{center}
\end{figure*}
\section{Appendix}
A few additional samples to showcase \texttt{EASTER}'s transcription performance on not so clear machine printed samples (figure \ref{fig: machinie_success}). Figure \ref{fig: hw_success} showcases additional samples of successful transcription for handwritten text.
\end{document}